\documentclass{article}

\usepackage{arxiv}

\usepackage[utf8]{inputenc} % allow utf-8 input
\usepackage[T1]{fontenc}    % use 8-bit T1 fonts
\usepackage{hyperref}       % hyperlinks
\usepackage{url}            % simple URL typesetting
\usepackage{booktabs}       % professional-quality tables
\usepackage{amsfonts}       % blackboard math symbols
\usepackage{nicefrac}       % compact symbols for 1/2, etc.
\usepackage{microtype}      % microtypography
\usepackage{lipsum}		% Can be removed after putting your text content
\usepackage{graphicx}
\usepackage{natbib}
\usepackage{doi}
\usepackage{multirow}

\usepackage{caption}
\usepackage{url,lineno,microtype,subcaption}
\usepackage[onehalfspacing]{setspace}

\usepackage{bm}
\usepackage[ruled,vlined,linesnumbered]{algorithm2e}

\usepackage{amsmath}

\usepackage{graphicx}
\graphicspath{{.}{./image/}}

\usepackage{xcolor}

\author{
Huanle Zhang\\
Department of Computer Science\\
University of California, Davis\\
\texttt{dtczhang@ucdavis.edu} \\
\And
Nicharee Wisuthiphaet \\
Department of Food Science and Technology\\
University of California, Davis\\
\texttt{nwis@ucdavis.edu} \\
\And 
Hemiao Cui \\
Department of Food Science and Technology\\
University of California, Davis\\
\texttt{hacui@ucdavis.edu} \\
\And 
Nitin Nitin \\
Department of Food Science and Technology\\
University of California, Davis\\
\texttt{nnitin@ucdavis.edu}
\And 
Xin Liu \\
Department of Computer Science\\
University of California, Davis\\
\texttt{xinliu@ucdavis.edu}
\And 
Qing Zhao \\
Department of Electrical and Computer Engineering\\
Cornell University\\
\texttt{qz16@cornell.edu}
}

\DeclareMathOperator*{\argmin}{arg\,min\ }
\DeclareMathOperator*{\argmax}{arg\,max\ }

\begin{document}
% \onecolumn
% \firstpage{1}

\title{Spectroscopy Approaches for Food Safety Applications: Improving Data Efficiency Using Active Learning and Semi-Supervised Learning}

\maketitle

\begin{abstract}

The past decade witnessed rapid development in the measurement and monitoring technologies for food science. Among these technologies, spectroscopy has been widely used for the analysis of food quality, safety, and nutritional properties. Due to the complexity of food systems and the lack of comprehensive predictive models, rapid and simple measurements to predict complex properties in food systems are largely missing. Machine Learning (ML) has shown great potential to improve the classification and prediction of these properties. However, the barriers to collecting large datasets for ML applications still persists. 
In this paper, we explore different approaches of data annotation and model training to improve data efficiency for ML applications.  Specifically, we leverage Active Learning (AL) and Semi-Supervised Learning (SSL) and investigate four approaches: baseline passive learning, AL, SSL, and a hybrid of AL and SSL. To evaluate these approaches, we collect two spectroscopy datasets: predicting plasma dosage and detecting foodborne pathogen. Our experimental results show that, compared to the \textit{de facto} passive learning approach, advanced approaches (AL, SSL, and the hybrid) can greatly reduce the number of labeled samples, with some cases decreasing the number of labeled samples by more than half.

% \tiny
%  \keyFont{ \section{Keywords:} food science, spectroscopy analysis, machine learning, data efficiency, active learning, semi-supervised learning} 
\end{abstract}

\section{Introduction}

\begin{figure}[!t]
    \centering
    \includegraphics[width=0.9\textwidth]{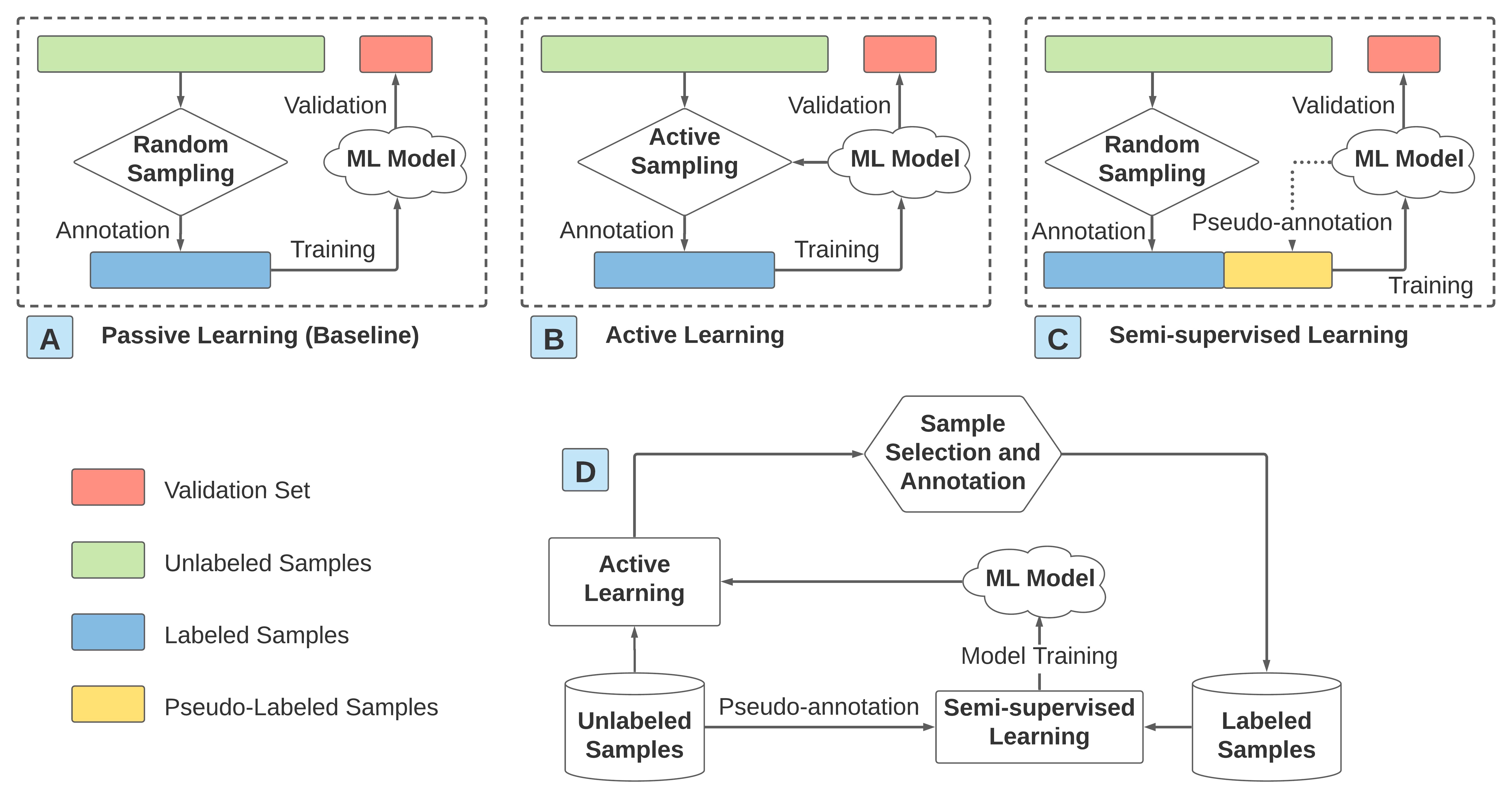}
    \caption{Four approaches of data annotation and model training. (A) Passive learning. (B) Active learning. (C) Semi-supervised learning. (D) Hybrid of active learning and semi-supervised learning. }
    \label{fig:scenario}
\end{figure}

Rapid measurement and monitoring technologies are being developed for diverse applications in food science. The goal of these technologies is to develop predictive relationships that can be used to better monitor and enhance the quality, safety and nutritional properties of food. Among these measurement approaches, spectroscopic analysis has been widely used to analyze food properties. 
To fully realize the great potential of these technologies, several key barriers need to be overcome before their transfer to industrial applications. 
%Despite the significant potential, there are key barriers in translating these technologies to industrial applications.
The discovery of predictive relationships between the measurements and properties of food systems is one of the key limitations. This limitation results from the complexity of food systems and the lack of comprehensive predictive models that can use rapid and simple measurements to predict complex properties in food systems. ML has shown significant potential to improve the classification and prediction of these properties. However, the barriers to collecting large datasets for ML applications persists.

With the advances in computational capabilities and big data technologies, ML has been applied to a variety of agriculture and food-related fields~\citep{ml_food, ml_agriculture, ml_water}. A prevailing approach is to train an ML model using labeled samples. While unlabeled samples can often be collected at relatively low cost, annotating each sample to create its label is expensive
%Before training an ML model, practitioners often need to collect labeled samples. Typically, it is convenient to collect unlabeled samples. In contrast, annotating samples is costly 
and time-consuming, as it often involves human inspection and in-field experiments. For example, to predict the vineyard yield, robot carrying cameras can be used to collect a large number of unlabeled images, but practitioners still need to manually process each image in order to assign appropriate labels to the images~\citep{vineyard_yield}. The prevailing practice is to randomly select samples and label them via costly in-field experiments and human annotation processes, and the ML model is trained only using the labeled samples. This approach results in poor data efficiency~\citep{burr_active_learning_book, semi_supervised_learning_book}.  

\subsection{AL and SSL for Data-Efficient Model Training}

To improve data efficiency, we explore two advanced model training techniques: AL~\citep{burr_active_learning_book} and SSL~\citep{semi_supervised_learning_book}. Instead of randomly selecting unlabeled samples for annotation, AL selects samples for annotation based on how informative these samples are to the currently trained ML model. In comparison, SSL exploits unlabeled samples by assigning pseudo-labels to them and trains the ML model using both labeled and pseudo-labeled samples.

In this paper, we study four approaches of data annotation and model training as illustrated in Fig.~\ref{fig:scenario}. (A) In the baseline passive learning approach, samples are randomly selected from the pool of unlabeled samples for annotation, and the ML model is trained only on the labeled samples. (B) In the AL approach~(details in Section~\ref{sect:active_learning}), the selection of unlabeled samples is dependent on the currently trained ML model. Specifically, AL selects the unlabeled samples most useful for training the ML model. Various sampling strategies are explored, which quantify the usefulness of unlabeled samples based on different criteria. The ML model is trained only using the labeled samples. 
(C) In the SSL approach (details in Section~\ref{sect:semi-supervised_learning}),  unlabeled samples are randomly selected for annotation. Instead of training the ML model using only the labeled samples, SSL assigns pseudo-labels to the  unlabeled samples. The ML model is trained using both the labeled and pseudo-labeled samples. Therefore, the number of training samples in SSL equals the total number of samples. The methods of assigning pseudo-labels can be either related to the currently trained ML model (e.g., self-training) or rely only on the relationship among samples (e.g., label spreading). 
(D) In the hybrid approach that integrates AL and SSL, AL selects an unlabeled sample for annotation, and SSL assigns pseudo-labels to the remaining unlabeled samples in each iteration. The ML model is trained using both labeled and pseudo-labeled samples. In this hybrid approach, AL and SSL interact with each other in the following manner. The sampling strategy in AL is dependent on the ML model, which is trained using labeled samples from human annotation and pseudo-labeled samples from SSL. Meanwhile, labeled samples from AL affect SSL regarding which samples are required for pseudo-annotation.

% (1) \emph{Plasma Dosage Classification}. As an emerging nonthermal processing technology, plasma is highly effective in inactivating various types of food pathogens in solution as well as on food contact surfaces~\citep{plasma_review}. The dataset for plasma dosage classification was obtained by subjecting substrate to plasma treatment at various dosage levels and characterizing biochemical changes of substrate with Fourier-Transform Infrared Spectroscopy (FTIR). The data comprise aborbance intensities at various IR wavenumbers of the substrate under plasma treatment.
%We use the plasma dosage classification dataset to train a multi-classification ML model. (2) \emph{Foodborne Pathogen Detection}. Detection of the bacterial foodborne pathogen is one of the critical processes in the food and agricultural industry as it ensures the safety of food products before distribution and also reduces the risk of foodborne illness outbreak~\citep{pathogen_review}.  The dataset for pathogen detection was obtained by using bacteriophage for infection and lysis of the target bacterial pathogen in food samples. The fluorescence spectroscopy was then used to acquire the fluorescence Excitation Emission Matrices (EEM)~\citep{t7phage20chemical} of each sample. The data comprise fluorescence emission intensities at several pairs of excitation/emission wavelengths. We use the pathogen detection dataset to train a binary classification ML model. 

To evaluate different approaches for spectroscopic analysis in the food science field, we collect two datasets: plasma dosage classification and foodborne pathogen detection.
Atmospheric plasma technologies are being developed as a non-thermal processing technology to improve food safety, reduce the impact on food quality, and improve the sustainability of food processing operations. One of the key challenges in plasma technology applications is the rapid assessment of its efficacy in the sanitation of food contact surfaces and food products. With this motivation, we are developing infra-red spectroscopy methods to aid in assessing the dosimetry of plasma treatment. Similarly, rapid detection of foodborne pathogens is a critical unmet need in food systems. In this direction, we are developing fluorescence spectroscopic methods based on molecular interactions between bacteriophages and their target bacteria to enable specific detection of bacteria.

We apply different ML models for the multi-class classification tasks. 
Representative methods are considered for AL and SSL. 
In the experiments, we adopt 5-fold cross-validation for both datasets. Our experimental results show that compared to the baseline passive learning approach, the numbers of labeled samples for the plasma dosage classification dataset and the pathogen detection dataset are reduced by more than $50\%$ using the hybrid approach. The promising results demonstrate that AL and SSL based approaches of data annotation and model training effectively improve data efficiency for spectroscopy-based ML applications.

\subsection{Related Work of Applying AL and SSL in Food Systems}

Both AL and SSL have been successfully applied to many domains, such as drug discovery~\citep{al_drug, ssl_drug}, material science~\citep{al_material, ssl_material}, and systems biology~\citep{opex20nature,ssl_biology}.
Research in food systems has adopted AL (a.k.a. optimal experimental design) to optimize non-ML model parameters, with the goal of reducing the number of experiments. It shows promising results of applying AL to applications such as determining partition coefficient in freeze concentration~\citep{freeze_al}, tuning micro-extraction in beer~\citep{beer_al}, identifying a rice drying model~\citep{rice_al}, and developing uniaxial expression of extracting oil from seeds~\citep{oil_al}. This paper differs from existing AL works in food science as we target general ML models while previous works are designed for specific and analytical models of few parameters. Similarly, SSL have successfully been used in food science problems such as determining tomato maturity~\citep{tomato_ssl} and tomato-juice freshness~\citep{tomato_juice_ssl}. To the best of our knowledge, this paper is the first work that extensively evaluates different AL and SSL approaches for the ML models with applications in food systems. 
% the food science community. 

\begin{table}[!t]
    \centering
    \begin{tabular}{c c}
        Notation & Meaning  \\\hline 
        \toprule
        $\mathcal{U}$ & the set of unlabeled samples \\
        $\mathcal{L}$ & the set of labeled samples \\
        $\theta$ & the currently trained ML model \\
        $\theta^+$ & an updated ML model by adding a new training sample \\
        $C$ & the number of classes for classification \\
        $x$ & a sample \\
        $y$ &  the label of the sample $x$ \\
        $x^*$ & the sample that has the maximum utility measure \\
        $y^*$ & the label of the sample $x^*$ \\
        $P_{\theta}(y|x)$ &  the probability that the sample $x$ belongs to the label $y$ under the model $\theta$ \\ $\hat{y}$  & the most likely label for the sample $x$, i.e., $\hat{y} = \argmax_y P_{\theta}(y|x)$ \\\bottomrule
    \end{tabular}
    \caption{Notations used in this paper. }
    \label{tab:notation}
\end{table}

\section{Materials and Methods}

In this section, we first provide preliminaries for AL in Section~\ref{sect:active_learning} and SSL in Section~\ref{sect:semi-supervised_learning}. Then, we explain our two datasets in detail in Section~\ref{sect:dataset}. Last, our experimental setup is presented in  Section~\ref{sect:setup}. Table \ref{tab:notation} tabulates the notations that are used in this paper for quick reference. 

\subsection{Active Learning: A Primer}
\label{sect:active_learning}

AL, also known as query learning/optimal experimental design, is a sub-field of machine learning, which studies ML models that improve with experience and training~\citep{burr_active_learning_book}.
Compared with passive learning, AL considers the ``usefulness'' of unlabeled samples for the current ML model. It strategically selects unlabeled samples for annotation to improve data efficiency for ML model training.

Algorithm \ref{alg:pool-based} shows the workflow of the AL-based data annotation. $\mathcal{U}$ and $\mathcal{L}$ is the set of unlabeled samples and labeled samples, respectively (line 1-2). The ML model, denoted by $\theta$, is trained on the labeled sample set $\mathcal{L}$ (line 4). The following unlabeled sample, denoted by $x^*$, is selected that has the highest utility measure according to the currently trained model $\theta$ (line 5). Then, the experiment is conducted for $x^*$ to obtain its ground-truth label, denoted by $y^*$ (line 6). Since the label for $x^*$ is obtained, $x^*$ is removed from the pool of unlabeled samples $\mathcal{U}$, and the sample $x^*$ along with its label $y^*$ is added to the set of labeled samples $\mathcal{L}$ (line 7). The process repeats until the trained model $\theta$ has a satisfactory accuracy or does not improve with more labeled samples.

\begin{algorithm}[!t]
\SetAlgoLined

$\mathcal{U}$: set of unlabeled samples $\{x^{(u)}\}_{u=1}^U$

$\mathcal{L}$: set of labeled samples $\{\langle x, y\rangle^{(l)}\}_{l=1}^L$

\For{$t = 1, 2, ...$ }{
    $\theta = \textbf{train}(\mathcal{L})$\\
    select $x^* \in \mathcal{U}$, the unlabeled sample of the highest \emph{utility measure} according to the model $\theta$\\
    conduct experiment for $x^*$, which obtains its label $y^*$ \\
    remove $x^*$ from $\mathcal{U}$, and add $\langle x^*, y^* \rangle$ to $\mathcal{L}$ 
}
\caption{Selection of unlabeled samples for annotation in active learning.}
\label{alg:pool-based}
\end{algorithm}

The utility measure is essential for AL algorithms. There are various utility measures to estimate the usefulness of an unlabeled sample to the ML model. They mostly leverage the probability of the current ML model classifying an unlabeled sample $x$ to class $y$, i.e., $P_{\theta}(y|x)$. We introduce two categories of utility measure methods of AL: uncertainty-based sampling~\citep{al_uncertainty} and minimizing expected errors~\citep{al_expected_errors}.

\subsubsection{Uncertainty Sampling based Active Learning}

The premise of uncertainty sampling is that we can avoid annotating samples that the ML model is confident about and focus instead on the unlabeled samples that confuse the ML model. Least confident and entropy are the most-used metrics for measuring the uncertainty of unlabeled samples.

\begin{itemize}
    \item \textit{Least Confident}. For an unlabeled sample $x$, we can apply the currently trained ML model on it, which outputs the probability of the sample belonging to class $y$, i.e., $P_{\theta}(y|x)$, where $y = 1, 2, ..., C$ and $C$ is the total number of classes for classification. Let's denote by $\hat{y}$, the class that is most likely for the unlabeled sample $x$, i.e., $\hat{y} = \argmax_y P_{\theta}(y|x)$. The confidence of the current model for the sample $x$ is thus $P_{\theta}(\hat{y}|x)$. The unlabeled sample with the least confidence is selected for annotation, that is,
\begin{equation}
    x_{LC}^* = \argmax_x 1 - P_{\theta}(\hat{y} | x)
    \label{equ:least_confident}
\end{equation}     

\item \textit{Entropy}. Entropy is an often-used metric for quantifying the uncertainty of a distribution. For an unlabeled sample $x$, its entropy is calculated as $-\sum_y P_{\theta}(y | x) \cdot \log P_{\theta}(y | x)$, which is over the distribution of classes $y$ for $x$. The entropy-based uncertainty sampling method selects the sample with the maximum entropy, i.e., 
\begin{equation}
    x^*_E  = \argmax_x -\sum_y P_{\theta}(y | x) \cdot \log P_{\theta}(y | x)
    \label{equ:entropy}
\end{equation} 
\end{itemize}

\subsubsection{Minimizing Expected Error based Active Learning}

This category of AL algorithms aims to select samples to directly increase the model accuracy without relying on the assumption between the sampling strategy and the model performance (e.g., the ML model can avoid annotating confident samples as in the uncertainty sampling). Since the ground-truth label of an unlabeled sample $x$ is not available without experiment and annotation, the model accuracy by adding the sample $x$ and its label to the training set is unknown. Nonetheless, we can estimate the expected model accuracy when the sample $x$ is added for training, thus selecting a sample that minimizes the expected error of the current ML model.

Assume that the unlabeled sample $x$ belongs to the class $y$. 
Let $\theta^+$ denote the updated ML model by adding the sample $x$ and its imaginary label $y$ to the training set. Then, the expected prediction error of the model $\theta^+$ can be estimated by applying $\theta^+$ to all unlabeled samples in $\mathcal{U}$, i.e., $\sum_{x' \in \  \mathcal{U}} 1 - P_{\theta^+}(\hat{y}|x')$, where $1 - P_{\theta^+}(\hat{y}|x')$ is the prediction error for the unlabeled sample $x'$. Since the probability that $x$ belongs to class $y$ is $P_\theta(y|x)$, the expected prediction error of selecting the unlabeled sample $x$ for annotation is $\sum_y P_{\theta}(y|x) \left[\ \sum_{x' \in \  \mathcal{U}} 1 - p_{\theta^+}(\hat{y}|x') \right]$. Correspondingly, Eq.~(\ref{equ:expected_prediction_error}) formulates the sampling strategy that minimizes the expected prediction error. 
\begin{equation}
    x_{EPR}^* = \argmin_x \sum_y P_{\theta}(y|x) \left[\ \sum_{x' \in \  \mathcal{U}} 1 - p_{\theta^+}(\hat{y}|x') \right]
    \label{equ:expected_prediction_error}
\end{equation}
 
An alternative to minimizing the expected prediction error is to minimize the expected log-loss error. Log-loss error is the \textit{de facto} loss function for training classification models. Therefore, minimizing the expected log-loss error has a strong connection with ML model training. By replacing the prediction error $1 - p_{\theta^+}(\hat{y}|x')$ in Eq.~(\ref{equ:expected_prediction_error}) with the log-loss error $-\sum_{y'} P_{\theta^+}(y' | x') \cdot \log P_{\theta^+}(y' | x')$,  Eq.~(\ref{equ:expected_logloss_error}) formulates the sampling strategy of minimizing the expected log-loss error. 
\begin{equation}
    x_{ELR}^* = \argmin_x -\sum_y P_{\theta}(y|x) \left[\ \sum_{x' \in \  \mathcal{U}} \sum_{y'} P_{\theta^+}(y' | x') \cdot \log P_{\theta^+}(y' | x') \right]
    \label{equ:expected_logloss_error}
\end{equation}

\subsection{Semi-Supervised Learning: A Primer}
\label{sect:semi-supervised_learning}

\begin{algorithm}[!t]
\SetAlgoLined
$\mathcal{U}$: set of unlabeled samples $\{x^{(u)}\}_{u=1}^U$ \\
$\mathcal{L}$: set of labeled samples $\{\langle x, y\rangle^{(l)}\}_{l=1}^L$

% \For{$t = 1, 2, ...$ }{
\While{$\mathcal{U} \text{ is not empty}$}{
    $\theta = \textbf{train}(\mathcal{L})$\\
    select  $x^* \in \mathcal{U}$, the most confident sample according to model $\theta$\\
    obtain the pseudo-label of $x^*$ by applying the current model $\theta$ to it, i.e., $\theta(x^*)$\\
    remove  $x^*$ from $\mathcal{U}$  and add  $\langle x^*, \theta(x^*) \rangle$ to $\mathcal{L}$
}
\caption{Self-training based semi-supervised learning.}
\label{alg:self_train}
\end{algorithm}

SSL can be applied to exploit unlabeled samples. Typically, an SSL algorithm converts unlabeled samples to pseudo-labeled samples and then fine-trains the ML model using both labeled and pseudo-labeled samples. SSL can be categorized into inductive methods and transductive methods~\citep{semi_supervised_learning_survey20ML}. We introduce  self-training~\citep{selfTraining} and label spreading~\citep{labelSpreading}, which are representative methods from these two categories. 

\subsubsection{Self-Training based Semi-Supervised Learning}

In the self-training based method, the currently trained ML model is applied to pseudo-annotate the unlabeled samples~\citep{selfTraining}. 
Self-training methods assume that the prediction of the current model tends to be correct, and thus the model can be further fine-trained by leveraging more training samples.

Algorithm \ref{alg:self_train} shows the workflow of self-training based SSL.
It differs from the AL workflow (Algorithm \ref{alg:pool-based}) in two main parts. 
(1) The unlabeled sample $x$ is pseudo-labeled by the model itself, i.e., $\theta(x^*)$, in self-training, whereas it is human-annotated, i.e., the ground-truth label $y^*$, in AL (line 6 in Algorithm \ref{alg:self_train} versus line 6 in Algorithm \ref{alg:pool-based}). (2) The most confident unlabeled sample is selected in self-training. In contrast, the most uncertain unlabeled sample is chosen in AL (line 5 in Algorithm \ref{alg:self_train} versus line 5 in Algorithm \ref{alg:pool-based}). Self-training selects the most confident unlabeled sample to mitigate the error propagation of pseudo-annotation since the model $\theta$ is consecutively trained on the pseudo-labeled samples. In contrast, AL selects the model's most confusing unlabeled sample for human annotation so that the model can correctly classify misleading samples. The process of unlabeled sample selection, pseudo-annotation, and model training repeats until all the unlabeled samples are pseudo-annotated (line 3). At that time, the model has been trained on all labeled and pseudo-labeled samples.

\subsubsection{Label Spreading based Semi-Supervised Learning}

Label spreading based SSL typically adopts a graph structure, where the vertices are the samples (both labeled and unlabeled) and edges exist between neighbor vertices~\citep{labelSpreading}. The edge weight between two vertices represents the similarity of the corresponding two samples. The goal of label spreading is to assign pseudo-labels to the vertices of unlabeled samples. The labels of $x_i$ and $x_j$ are likely to be the same if the edge weight $w_{ij}$ between them is large. There are two mainstream methods to define the graph structure and the edge weights:
\begin{itemize}
    \item \emph{Fully Connected Graph}~\citep{ssl_graph}. In this graph, all vertices are connected with other vertices, and the edge weight $w_{ij}$ between $x_i$ and $x_j$ is calculated as 
\begin{equation}
    w_{ij} = \exp{(-\frac{||x_i - x_j||^2}{2\sigma^2})}
\end{equation}
where $||x_i - x_j||$ is the Euclidean distance between sample $x_i$ and sample $x_j$, and $\sigma$ controls the decreasing rate of the weight over the distance. The edge weight  $w_{ij}$ is 1 when $x_i = x_j$, and approximately 0 when $x_i$ is far from $x_j$. This weight representation is also called Radial Base Function (RBF) kernel or Gaussian kernel.

\item \emph{k-Nearest-Neighbor (kNN) Graph}~\citep{ssl_graph}. In the kNN graph, each vertex is connected to its $k$ nearest vertices. If sample $x_i$ and sample $x_j$ are connected, the edge weight $w_{ij}$ is 1; otherwise, $w_{ij}$ is 0. The kNN graph automatically adapts to the density of samples: in dense regions, the radius of the kNN neighborhood is small, while in sparse regions, the radius is large. 

\end{itemize}

The probability of the label for each unlabeled sample is obtained once the graph converges~\citep{labelSPread}. Then, each unlabeled sample is assigned to the label of the highest probability at once. Afterward, the ML model is re-trained using all samples (labeled and pseudo-labeled).

\subsection{Dataset}
\label{sect:dataset}

We collect two spectroscopy datasets in food science. One dataset predicts the plasma dosage, and the other detects the foodborne pathogen. Our datasets have representative data structures (1D and 2D spectroscopy), and we use different ML models for these two datasets.

\begin{figure}
    \centering
    \includegraphics[width=0.8\textwidth]{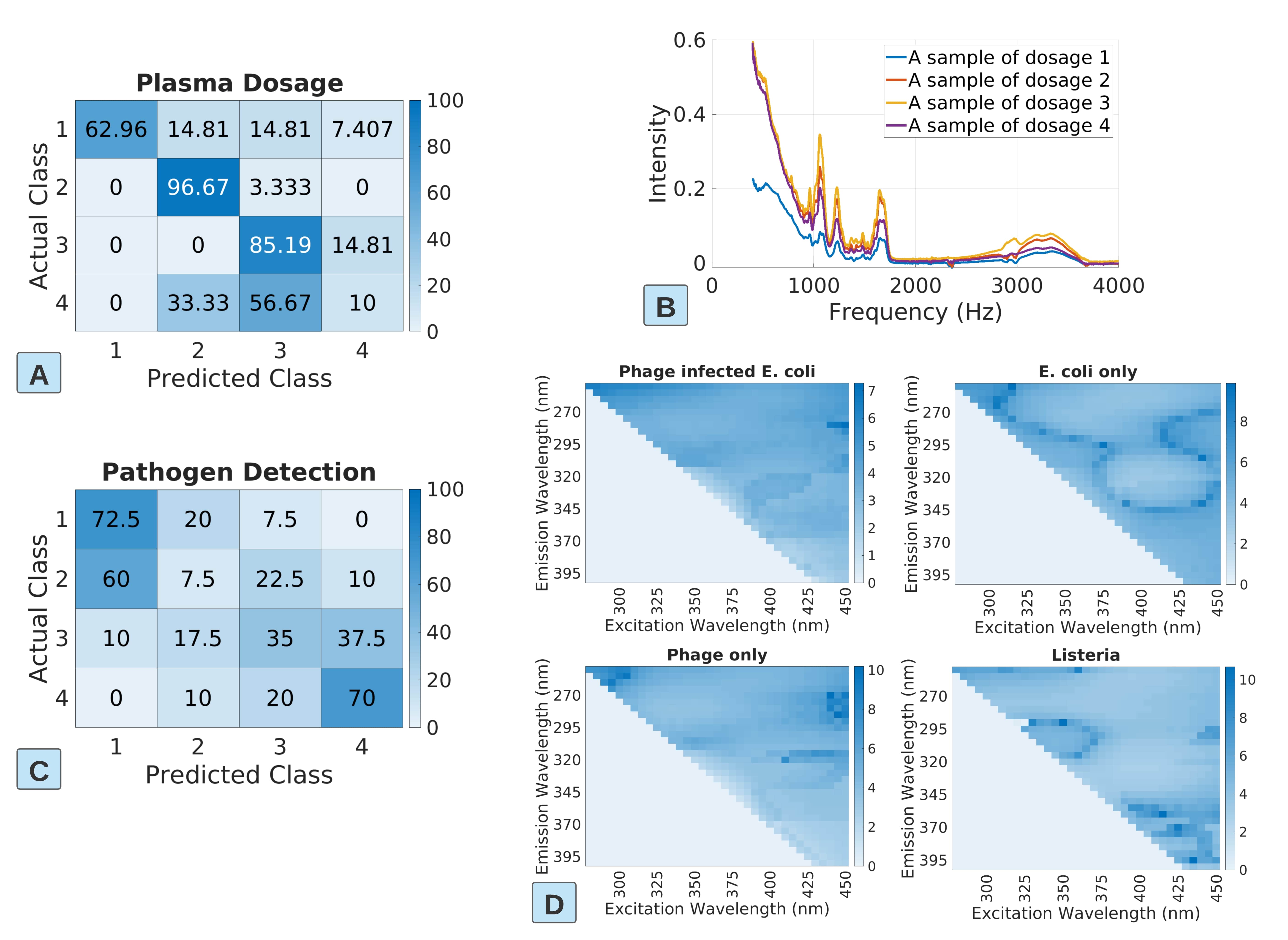}
    \caption{Two datasets are used in our experiments. The top row shows the plasma dosage classification dataset where (A) is the clustering result and (B) illustrates a sample for each class. The bottom row shows the pathogen detection dataset where (C) is the clustering result and (D) depicts a sample for each class. }
    \label{fig:dataset}
\end{figure}

\subsubsection{Dataset 1: Plasma Dosage Classification}

As an emerging nonthermal processing technology, plasma is highly effective in inactivating various types of food pathogens in solutions as well as on food contact surfaces~\citep{plasma_review}. One of the main goals of our plasma dosage classification dataset is to evaluate whether the plasma dosage can be predicated using the Fourier-Transform Infrared Spectroscopy (FTIR) spectral response of the DNA sample exposed to the plasma treatment. The dataset was obtained by subjecting the substrate to plasma treatment at various dosage levels and characterizing biochemical changes of the substrate with FTIR. The data comprise absorbance intensities at various IR wavenumbers of the substrate under plasma treatment.

Our 1D dosage classification dataset categorizes the plasma dosage into four classes, where classes 1, 2, 3, and 4 represent plasma injection of 0, 2, 5, and 10 minutes, respectively. In total, we collect 114 samples, where classes 1, 2, 3, and 4 have 27, 27, 30, 30 samples, respectively. To study the sample distribution over the feature domain, we apply K-Means~\citep{k_means} to group samples into 4 clusters. Fig.~\ref{fig:dataset}(A) shows the clustering result. It indicates that the class 2 and the class 3 samples are well clustered, while many of the class 4 samples are located within cluster 2 and cluster 3 (a PCA visualization is given in Fig.~\ref{fig:sample_selection}). Fig.~\ref{fig:dataset}(B) illustrates a sample for each class. Each sample contains 1868 values representing the intensity of the IR frequency reflectance in the range of 400 Hz to 4000 Hz with a step of 2 Hz.

\subsubsection{Dataset 2: Foodborne Pathogen Detection}

Detection of the bacterial foodborne pathogen is one of the critical processes in the food and agricultural industry~\citep{pathogen_review}. It ensures the safety of food products before distribution and reduces the risk of a foodborne illness outbreak. Among various types of bacteria, \textit{E. coli} has been used as an indicator of the fecal contamination and poor sanitary quality of food and water~\citep{t7phage83aem}. 
T7 bacteriophage or T7 phage has been used as a tool for \textit{E. coli} detection as it infects explicitly \textit{E. coli} cells results in bacteria cell lysis and release of the cell components along with the amplified T7 phage progenies~\citep{t7phage20plosone}. Therefore, bacterial cell lysis and T7 phage amplification can indicate \textit{E. coli} contamination in the samples. Fluorescence EEM spectroscopy is an analytical technique providing multi-dimensional information~\citep{t7phage20chemical} that has been used to detect organic materials, including bacterial cells~\citep{eem20water}.

We explore whether 2D EEM spectroscopy can detect the change of the physical and chemical properties of the samples due to the phage infection of \textit{E. coli}. We use classes 1, 2, 3, and 4 to represent phage infected \textit{E. coli}, \textit{E. coli} only, phage only, and \textit{Listeria} solutions, respectively. 
The EEM spectra of the resuspended samples were collected with the wavelength range of 250–400 nm for excitation and 260–450 nm for emission with 5-nm increments.
Fig.~\ref{fig:dataset}(C) shows the result of grouping the samples into 4 clusters using K-Means. It indicates that most \textit{E. coli} only and the phage only samples are wrongly categorized. Fig.~\ref{fig:dataset}(D) visualizes one sample for each class. In the heat maps, we take the logarithm of the frequency intensity at each frequency pair of the excitation and emission wavelength. In total, we collect 160 samples, and each class has 40 samples. Each sample includes 744 numbers representing the fluorescence response for the excitation-emission pairs. 

% \subsubsection{Dataset Summary}

% \huanle{move to Section 2.3}

% We select two spectroscopy datasets for experiments because they are commonly used in the food and agriculture sectors for monitoring quality, safety, and production of food. Our datasets are diverse and representative. The plasma dosage classification dataset is 1-D input, while the foodborne pathogen detection dataset is 2-D input. In addition, they target different tasks: one for multi-class classification and the other one for binary classification, and we apply different ML models for them. 

\subsection{Experimental Setup}
\label{sect:setup}

Given the same number of labeled samples, we compare the ML model accuracy when different data annotation and model training approaches are applied. 
We use the LightGBM multi-class classification model~\citep{lightgbm17nips} for the plasma dosage dataset, and the logistic regression classifier~\citep{logistic_regression} for the pathogen detection dataset. We select these ML models because they show promising results in our related projects. We adopt 5-fold cross-validation for both datasets. Specifically, at each cross-validation round, the training and validation sets comprise the four-fold samples and the remaining one-fold samples, respectively. We use the training set for annotating a given number of samples and apply the trained ML model to predict the validation set.
Note that the training set is assumed to be unlabeled at the beginning of each cross-validation, and the model is trained with different approaches~(refer to Fig.~\ref{fig:scenario}). 
The predictions for all samples are obtained by aggregating the predictions of the ML model for each validation set from the five cross-validations.

To determine the hyper-parameters (e.g., learning rate, regularization) of ML models for a given number of labeled samples, we apply Optuna~\citep{optuna} to the passive learning approach and adopt the same hyper-parameters for all approaches. Therefore, the hyper-parameters favor the passive learning approach in our experiments.  Using Optuna, we only need to provide reasonable ranges for hyper-parameters, and Optuna can identify the optimal hyper-parameters. We find that hyper-parameters from Optuna tend to have higher model accuracy than the default hyper-parameters.

\section{Results}

\begin{figure}[!t]
    \centering
    \centerline{
    \includegraphics[width=1\textwidth]{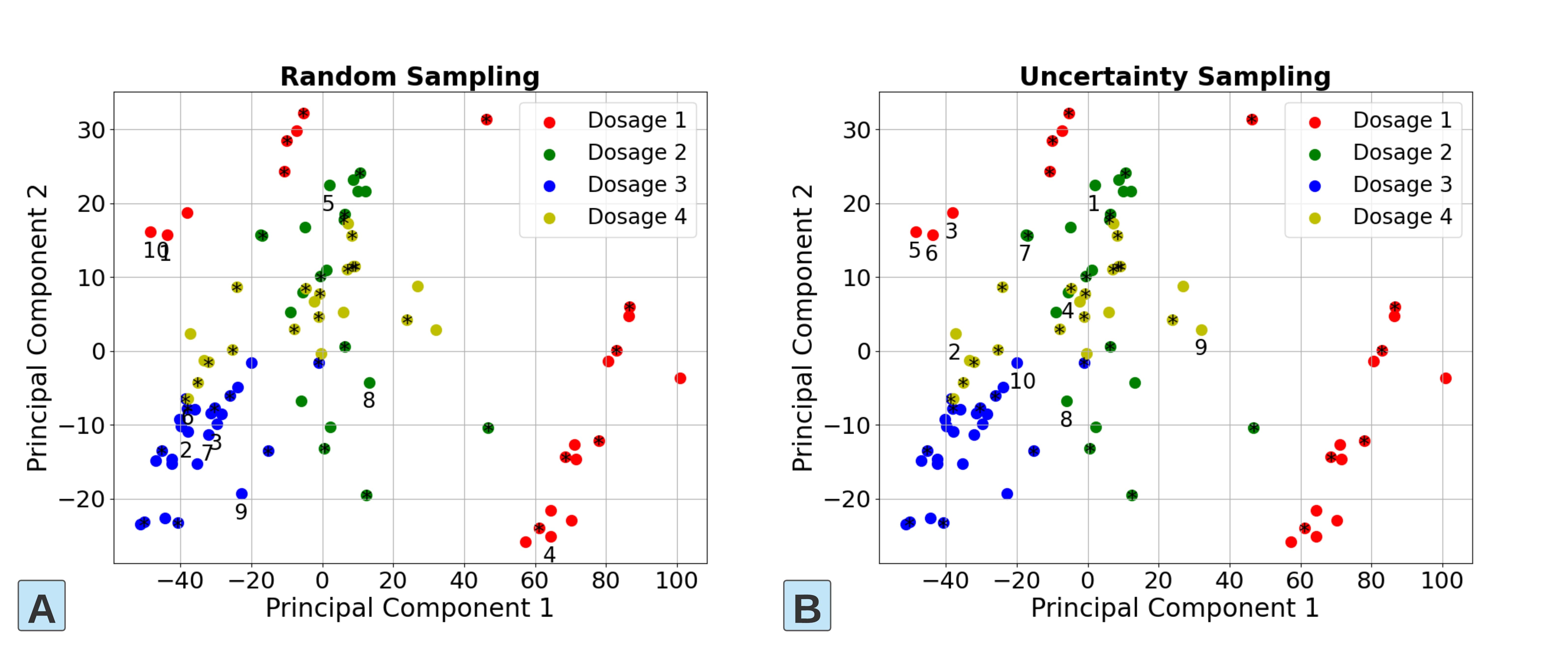}
    }
    \caption{Demonstration of the samples selected in the passive learning approach and the active learning approach for the plasma dosage classification dataset. (A) Passive learning approach. (B) Active learning approach with the entropy-based uncertainty sampling. The 40 randomly selected samples for training an initial ML model are marked with ``*''. The next 10 samples are tagged with the order number.}
    \label{fig:sample_selection}
\end{figure}

We evaluate the four approaches of data annotation and model training using our two datasets. We warm-start a LightGBM model using 40 randomly selected labeled samples for the plasma dosage dataset and then apply different approaches.  Likewise, we first train a logistic regression classification model using 25 randomly selected labeled samples for the pathogen detection dataset before applying different approaches. We run 10 experiments to average the results. Since we adopt the 5-fold cross-validation, the maximum number of samples for training in each cross-validation is 90 for the plasma dosage task and 127 for the pathogen detection task.

\subsection{Results of the Active Learning Approach}

We first demonstrate the order of samples selected for annotation in the passive learning and AL approaches. Principal Component Analysis (PCA)~\citep{pca} is applied to project the high-dimensional samples into two components for visualization. Fig.~\ref{fig:sample_selection}
illustrates a trace for the passive learning approach and the AL approach for the dosage classification dataset, respectively. 
As Fig.~\ref{fig:sample_selection}(A) illustrates, the selected samples in the passive learning can be close to each other and also near the center of the class~(e.g., samples 2, 3, and 7), which are less likely to improve the ML model accuracy. In comparison, as Fig.~\ref{fig:sample_selection}(B) shows, the selected samples in the AL approach are close to the boundaries of different classes. In these boundary areas, the ML model is difficult to classify. Thus, training with the samples in the boundary areas is more likely to increase the ML model accuracy.

\begin{figure}[!t]
    \centering
    \centerline{
    \includegraphics[width=0.9\textwidth]{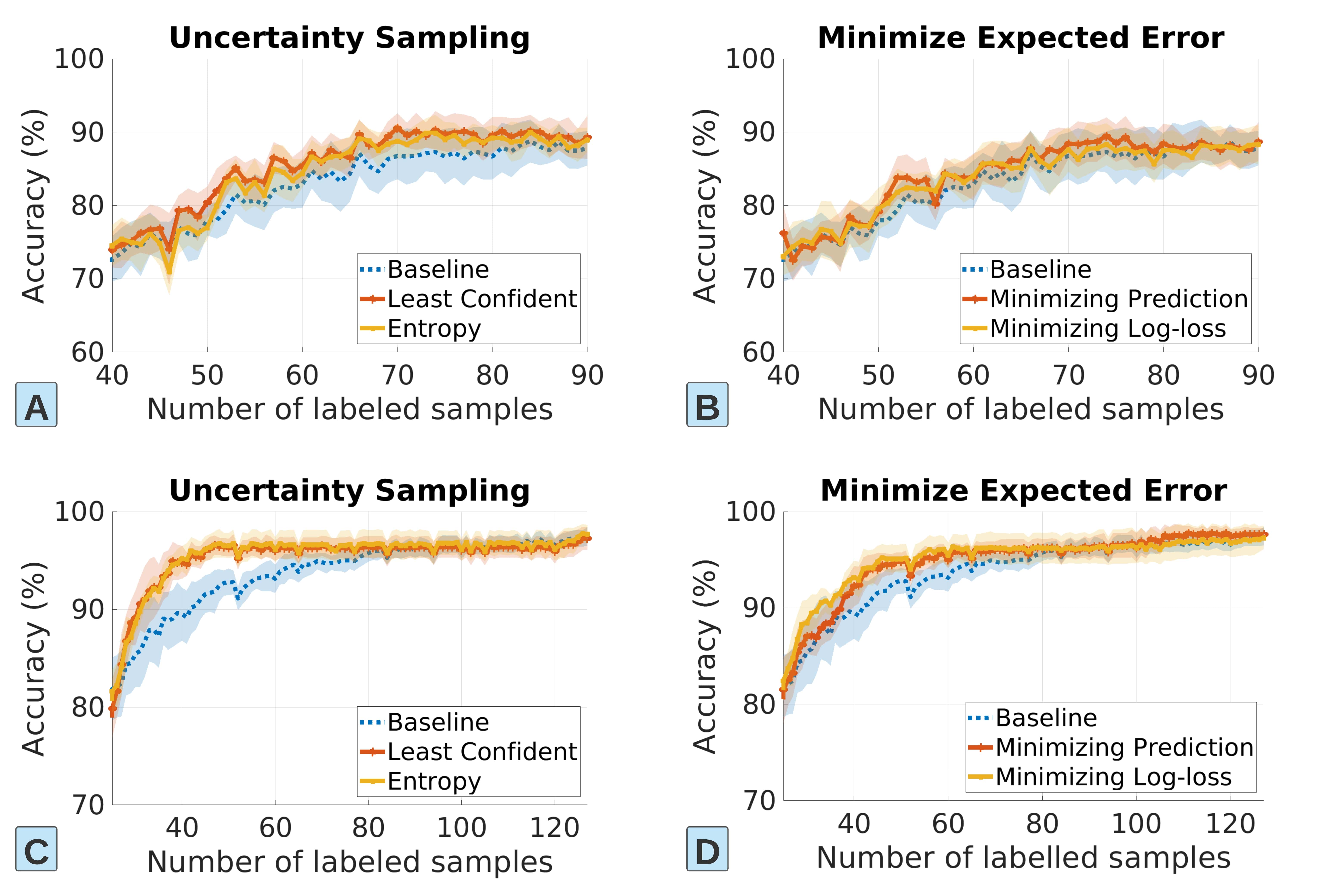}
    }
    \caption{The ML model performance versus the number of labeled samples in the AL approach. The first row and the second row show the results for the dosage classification dataset and the pathogen detection dataset, respectively. (A) and (C): uncertainty sampling based on least confident and entropy; (B) and (D) minimizing expected prediction error and expected log-loss error based sampling.}
    \label{fig:pool_active_learning}
\end{figure}

In the rest of the experiments for the AL approach, we consider (1) uncertainty sampling methods based on least confident and maximum entropy, and (2) minimizing the expected prediction error and the expected log-loss error.
Fig.~\ref{fig:pool_active_learning} plots the ML model accuracy versus the number of labeled samples. The first row and the second row show the model performance for the dosage classification and the pathogen detection datasets, respectively. We show the standard deviation in shaded colors.  We can see that AL approach achieves better performance than the passive learning approach, especially for the foodborne pathogen dataset. For example, the uncertainty-based AL approach reduces the number of labeled to only 40 for the pathogen detection dataset, compared to 80 in the passive learning approach, achieving 50\% label data reduction.  The sampling strategies of minimizing expected errors depend on the current ML model accuracy to calculate the expected errors, which does not perform well if the current ML model has low prediction accuracy (e.g., for the plasma dosage classification dataset). Nonetheless, we do not see degradation of data efficiency.

\subsection{Results of the Semi-supervised Learning Approach}

\begin{figure}[!t]
    \centering
    \centerline{
    \includegraphics[width=0.9\textwidth]{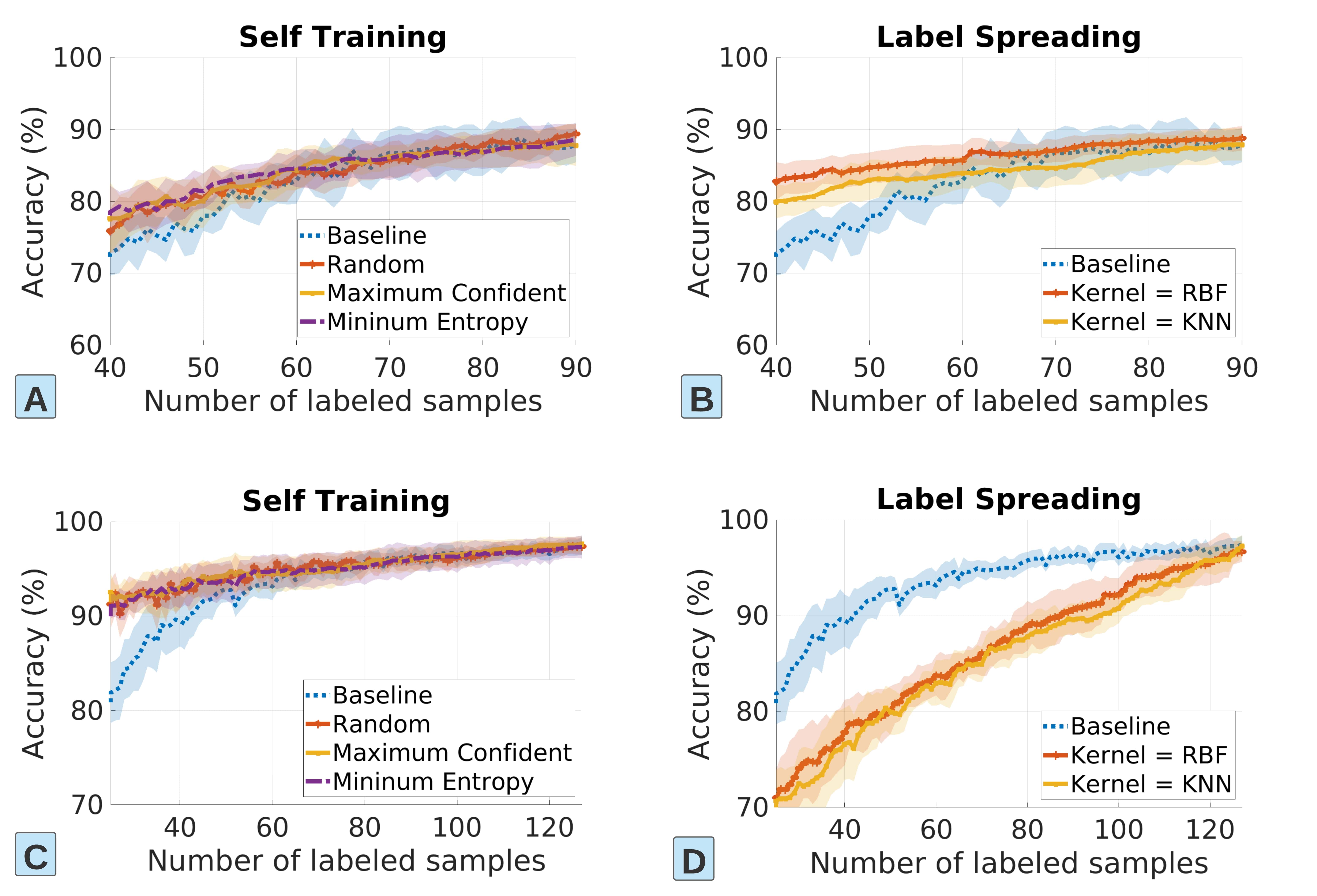}
    }
    \caption{The ML model accuracy versus the number of labeled samples in the semi-supervised learning approach. The first row and the second row show the results for the dosage classification dataset and the pathogen detection dataset, respectively. (A) and (C): self-training based on maximum confident and minimum entropy; (B) and (D): label spreading based on kNN kernel and RBF kernel.}
    \label{fig:semi_supervised_learning}
\end{figure}

SSL exploits unlabeled samples by assigning pseudo-labels to them and then trains the ML model using both labeled and pseudo-labeled samples. In the evaluation of the SSL approach, we consider self-training and label spreading. We set the number of neighbors in the kNN kernel to 7 and the $\sigma$ in the RBF kernel to 0.1 for both datasets.

Fig.~\ref{fig:semi_supervised_learning} plots the ML model accuracy versus the number of labeled samples in the SSL approach. The first row and the second row show the model performance for the plasma and pathogen datasets. We visualize the standard deviation in shaded colors. We can see that the SSL approach can improve data efficiency in most experiment scenarios. The exception is the label-spreading for the pathogen detection dataset, which shows much-degraded performance. It is probably because the pathogen detection dataset is poorly clustered (see Fig.~\ref{fig:dataset}(C)), and thus the pseudo-labels from label-spreading are mostly incorrect. In the cases of label spreading for the plasma dataset and self-training for the pathogen dataset, SSL approach improves the model accuracy by about 10\% when the label samples are few.

\subsection{Results of the Hybrid Approach}

\begin{figure}[!t]
    \centering
    \centerline{
    \includegraphics[width=0.9\textwidth]{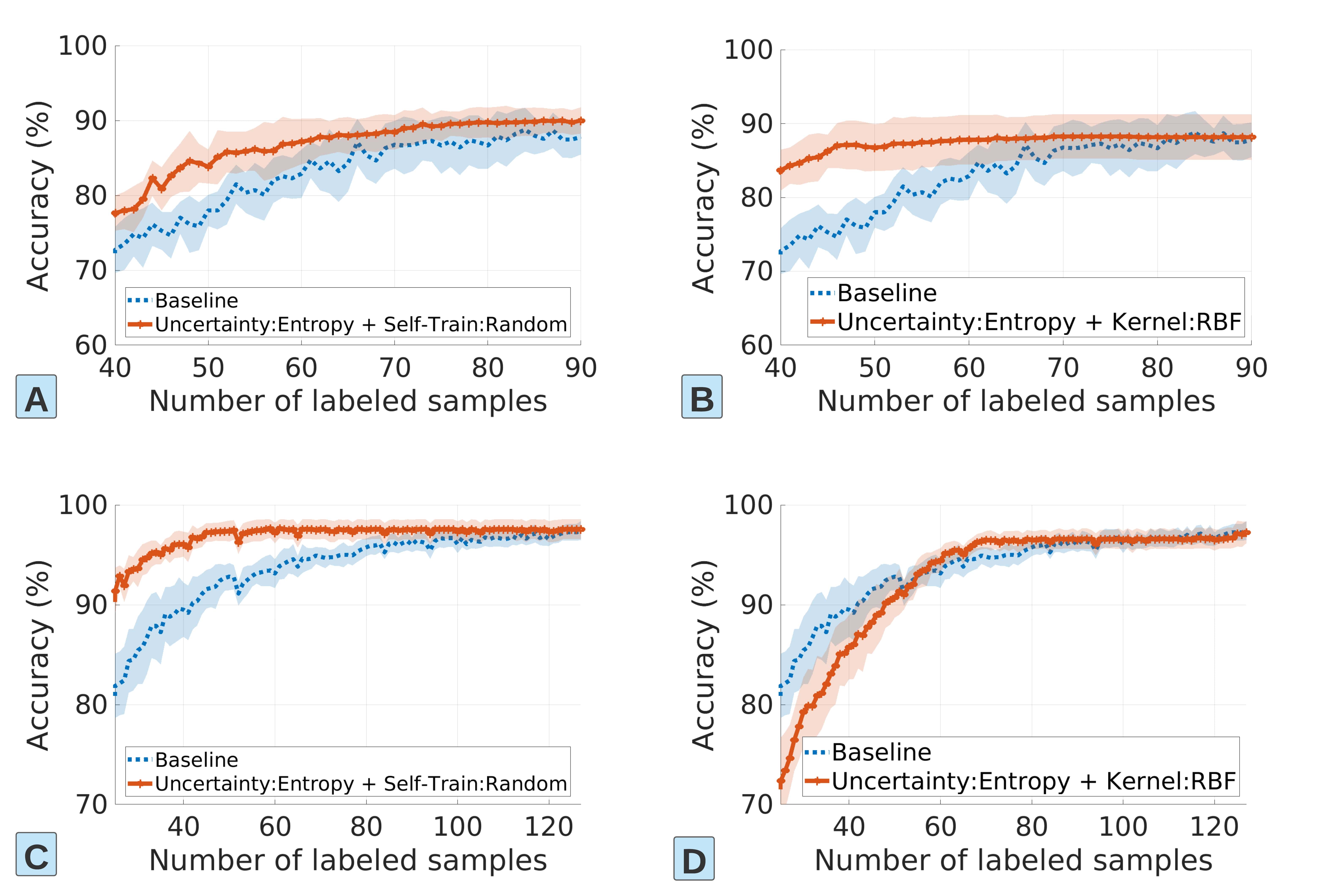}
    }
    \caption{The ML model accuracy versus the number of labeled samples in the hybrid approach. The top row and the bottom row show the results for the plasma dosage dataset and the foodborne pathogen detection dataset. (A) and (C) Uncertainty-based AL with self-training based SSL. (B) and (D) Uncertainty-based AL with self-labeling based SSL. }
    \label{fig:combine}
\end{figure}

We measure the performance of the hybrid approach using two combinations: random sampling based self-training and RBF-based self-labeling for SSL, both with entropy-based uncertainty sampling for AL. The $\sigma$ value in the label spreading is set to 0.01 for both datasets.

Fig.~\ref{fig:combine} shows the ML model accuracy for the hybrid approach, where the top row is for the plasma dosage dataset and the bottom row is for the pathogen detection dataset. As we can see, the hybrid approach dramatically improves the data efficiency for ML model training. For example, Fig.~\ref{fig:combine}(B) shows that the hybrid approach reaches the maximum model accuracy with only 45 labeled samples, in comparison to 85 in the baseline, and thus reduces the amount of labeled data by about 50\%. The pathogen detection dataset performs even better with the hybrid approach: 50 labeled samples achieve the same maximum model accuracy as 120 labeled samples in the baseline, reducing the number of labeled samples by about 60\%. Even though the self-labeling based SSL works poorly for the pathogen dataset (Fig.~\ref{fig:semi_supervised_learning}(D)), the hybrid approach still works better than the baseline: 70 labeled samples vs. 90 labeled samples. Overall, the hybrid approach is promising in improving data efficiency.

%Fig.~\ref{fig:combine}(A) illustrates that, compared to the AL approach (Fig.~\ref{fig:pool_active_learning}(A)(B)) and the SSL approach (Fig.~\ref{fig:semi_supervised_learning}(A)(B)), the hybrid approach for the dosage classification dataset further improves data efficiency for the ML model training. Specifically, $\sim\!45$ labeled samples achieve the same accuracy of using all 90 labeled samples, reducing the number of labeled samples by $50\%$. However, for the pathogen detection dataset, the hybrid approach has approximately the same performance as the passive learning approach (Fig.~\ref{fig:combine}(B)). As we previously discussed, the pathogen detection dataset is not well-clustered, and thus SSL does not work well. The benefit of AL is suppressed by integrating it with SSL, indicating the interaction between AL and SSL is inherently complicated. 

%From the results of the hybrid approach, we make the following observations. (1) The hybrid approach can provide supreme performance, which greatly improves the data efficiency of ML model training (e.g., the dosage classification dataset). (2) The hybrid approach does not guarantee better performance than the AL approach and the SSL approach. For the pathogen detection database, the AL approach works better than the hybrid approach. Nevertheless, we do not see a noticeable model performance degradation using the hybrid approach compared to the passive learning approach. 

\section{Discussion}

We explore different approaches of data annotation and model training to improve data efficiency for ML applications. In this section, we discuss practical considerations in applying these advanced approaches.  

\subsection{Practicability of Advanced Approaches}

The passive learning approach has been adopted for the majority of projects. An essential question is whether more advanced approaches can be effectively and efficiently applied for new projects, considering the overheads of implementing these approaches. We argue that when the annotation process is costly, the benefits of the reduced number of labeled samples overweigh the implementation overheads. In fact, AL and SSL have been successfully applied to many domains~\citep{ssl_drug,al_material, ssl_material,opex20nature,ssl_biology}. For example,  the medicine industry applies AL to discover antagonists for dopamine $D_4$~\citep{drgu_d4} and CXC-chemokine~\citep{drgu_chemokine}, among the estimated range of $10^{30}$-$10^{60}$ drug-like molecules~\citep{al_drug}. In food systems, development of AL and SSL methods can help address challenges of obtaining labeled samples for ML models as labeling in many food safety applications is time-consuming and labor-intensive. 

% many of these labeled samples in food safety applications require both resources and extended time. 

\subsection{Determining the Optimal Approach}

We introduce three advanced data annotation and model training approaches and compare their performance with the passive learning approach. Our evaluation results show that the hybrid approach can dramatically improve data efficiency. Therefore, we suggest the hybrid approach that leverages both AL and SSL. However, the sampling strategy in AL and the pseudo-labeling method in SSL are crucial for the hybrid approach's performance.
%, which requires careful consideration. 

%the passive learning approach performs the worst compared to the other three approaches. However, there is no guarantee on which approach works best \textit{a priori}. For example, the hybrid approach works exceptionally well for the dosage classification dataset, while the AL approach works the best for the pathogen detection dataset. Nonetheless, there are several rules of thumb drawn from the community~\citep{burr_active_learning_book,semi_supervised_learning_survey20ML}. 
%Based on our experience, we suggest the following approach selection procedure: (1) If the data samples are poorly clustered, SSL is not expected to work. In this case, we advocate the AL approach. To decide whether data samples can be clustered, we can either apply PCA visualization or k-means clustering algorithms~\citep{k_in_k_means} to check whether the number of clusters is close to the number of classes for classification. (2) If the data samples are well-clustered, we suggest the hybrid approach that leverages both AL and SSL. 

\subsection{Determining the Optimal Sampling Strategy in Active Learning}

There are a great variety of sampling strategies available in AL~\citep{burr_active_learning_book}. 
Although it is not possible to obtain a universally good AL sampling strategy~\citep{greedyactive04nips}, we have demonstrated that by using simple sampling strategies such as uncertainty-based sampling, we achieve better data efficiency than the passive learning approach. Simple sampling strategies based on uncertainty and disagreement are recommended if the domain knowledge about the samples and the problems are not available~\citep{burr_active_learning_book}. On the other hand, incorporating domain knowledge into AL can further improve data efficiency. For example, \cite{alice20emnlp} proposes an expert-in-the-loop AL framework that utilizes language explanations from domain experts to iteratively distinguish misleading breeds of birds, which outperform baseline models that are trained with 40\%-100\% more training samples. Automatic selection of sampling strategies has also been extensively studied, e.g., \citep{al_learning_15aaai} and~\citep{alFromData17nips}.

In our experiments, we warm-start the initial ML model by randomly selecting and training 40 samples for the dosage classification dataset and 25 samples for the pathogen detection dataset. Another common practice is to switch between random sampling (for exploration) and AL sampling strategy (for exploitation). For example, \cite{opex20nature} train a predictive ML model of gene expression with 44\% fewer data by consecutive switching between random sampling and mutual information based AL sampling strategy.

\subsection{Determining the Optimal Pseudo-Labeling Method in Semi-Supervised Learning}

In addition to the self-training and the label-spreading methods, semi-supervised learning includes many other methods such as co-training, boosting, and perturbation-based~\citep{semi_supervised_learning_survey20ML}. Similar to AL, SSL does not guarantee better data efficiency than the passive learning approach~\citep{hurt_ssl}. Nonetheless, our experiments demonstrate that simple pseudo-labeling methods such as self-training can greatly improve the model accuracy when the underlying samples can be well-clustered and the number of labeled samples are small. 
Recent advances in SSL show promising results of the perturbation-based semi-supervised neural networks, which empirically and consistently outperforms the passive learning approach~\citep{semi_supervised_learning_survey20ML}. We leave it as future work to evaluate the perturbation-based methods.

\subsection{Complexity Analysis}

\begin{table}[!t]
    \centering
    \centerline{
\scalebox{0.9}{
    \begin{tabular}{c c c c}
        \textbf{Approach} & \textbf{Method} & \textbf{\# of Model Training} & \textbf{\# of Model Inference} \\[2pt]\toprule
        \\[-5pt]
        Passive & - & $F - B$ & 0 \\[3pt]\hline
        \\[-5pt]
         \multirow{2}{*}{AL} & Uncertainty-based & $F - B$ & $\frac{(F-B)\times (2T - F - B + 1)}{2}$ \\
         \\[-5pt]
          & Minimize expected errors & $\frac{C\times (F - B) \times (2T - F - B + 1)}{2}$ & $\approx\! \frac{3C \times (F-B)^2 \times (T - B)^2 + (F - B)^3}{3}$ \\[5pt]\hline
          \\[-5pt]
         \multirow{2}{*}{SSL} & Self-training & $\frac{2\times (F - B) + (T-B)\times (T-B+1)}{2}$ & $\frac{(T - B) \times (T - B + 1)}{2}$ \\
         \\[-5pt]
          & Label-spreading & $F - B$ & 0 \\[2pt]\hline 
          \\[-5pt]
          \multirow{2}{*}{Hybrid} & Uncertainty + Self-training & $\frac{2\times (F - B) + (T-B)\times (T-B+1)}{2}$ & $\frac{(F-B)\times (2T - F - B + 1) + (T - B) \times (T - B + 1)}{2}$ \\
          \\[-5pt]
           & Uncertainty + Label-spreading & $ F - B$ & $\frac{(F-B)\times (2T - F - B + 1)}{2}$ \\[2pt]\bottomrule 
    \end{tabular}
}
    }
    \caption{Computation overheads of different approaches of data annotation and model training. $C$ is the number of classes for classification; $T$ is the total number of samples, $B$ is the initial labeled samples, and $F$ is the number of final labeled labels. }
    \label{tab:complexity}
\end{table}

The computation capability is also a factor in deciding the approach of data annotation and model training. We formulate the computation complexity as the number of ML model training and ML model inference, as they are more computation demanding than other operations (e.g., graph construction in label-spreading). We denote the total number of samples by $T$, the number of the initial labeled samples by $B$, and the number of final labeled samples by $F$. Table~\ref{tab:complexity} tabulates the overheads in different approaches, whereas we ignore the model validation overheads as they are the same for all approaches. 
If the model is heavy and slow to execute, then the computation-intensive methods, e.g., minimizing expected errors, should be avoided.

\subsection{Extension to Regression Problems}

%Both classification and regression are common problems. 
In this paper, we use classification to illustrate different approaches of data annotation and model training.  
%Although methods designed for classification may not be applied to regression problems, 
We expect our main conclusion to remain valid for regression problems; that is, advanced approaches can help improve data efficiency for model training compared to the passive learning approach. There are many existing works solving regression problems in AL and/or SSL~\citep{al_regression,ssl_regression}, which we will explore in our future work. 

\subsection{Applicability for Diverse Applications in Food Quality, Traceability, and Safety}
% Spectroscopic analysis has been widely used for the analysis of food quality, safety, and nutrition in food science. We collect two spectroscopy datasets in our experiments, which show promising results of using advanced approaches. We believe that AL and SSL can also improve data efficiency for other food science problems. .........

The spectroscopy measurement approaches selected for this study represent two distinct spectral methods, namely IR spectroscopy and fluorescence spectroscopy. IR spectroscopy has been proposed for diverse applications in food quality~\citep{food_quality}, traceability~\citep{food_traceability} and food safety~\citep{food_safety}. These diversity of applications are enabled by the unique ability of IR spectroscopy to detect molecular composition of food materials using non-destructive sampling and rapid data collection. ML approaches can complement the current chemometric methods used for the analysis of IR spectroscopy data. Complementary to IR spectroscopy, fluorescence spectroscopy approaches rely on the photo-active properties of fluorophores in food systems and their relationship with changes in food quality and traceability~\citep{fluorescence}. Similarly, fluorescence properties have also been used to monitor the presence of bacteria in water samples~\citep{fluorescence_water}. However, due to significant interference between the food materials and bacterial components there has been limited applications of fluorescence spectroscopy in food safety. To address some of these limitations, this study evaluated the applications of Fluorescence EEM spectroscopy for food safety. EEM is a technique that allows for the complete, quantitative determination of the fluorescence profile of a given material and has been used for biomedical and material characterization applications~\citep{fluorescence_biomedical}.  Application of EEM spectroscopy including their integration with ML methods can address some of the key challenges in  application of fluorescence spectroscopy for food applications.

\section{Conclusion}

In this paper, we target
data efficiency of ML applications for spectroscopy analysis in food science.  To mitigate the annotation cost by reducing the number of labeled samples, we explore different approaches of data annotation and model training: passive learning, active learning, semi-supervised learning, and the hybrid of active learning and semi-supervised learning. We evaluate these approaches in two spectroscopy datasets and find that advanced approaches can greatly improve data efficiency for ML model training. These approaches are general and thus can be applied to various ML-based food science research.

% \input{abstract}

% \input{introduction}

% \input{materials_and_methods}

% \input{results}

% \input{discussion}

% \input{conclusion}

% \section{Additional Requirements}

% For additional requirements for specific article types and further information please refer to \href{http://www.frontiersin.org/about/AuthorGuidelines#AdditionalRequirements}{Author Guidelines}.

\section*{Conflict of Interest Statement}
%All financial, commercial or other relationships that might be perceived by the academic community as representing a potential conflict of interest must be disclosed. If no such relationship exists, authors will be asked to confirm the following statement: 

The authors declare that the research was conducted without any commercial or financial relationships that could be construed as a potential conflict of interest.

\section*{Author Contributions}

NN and XL conceptualized and supervised the project. QZ supervised the project. HZ conceptualized and implemented the project. NW collected the pathogen detection dataset and conducted a preliminary machine learning model evaluation. HC collected the plasma dosage classification dataset and conducted a preliminary machine learning model evaluation. All authors contributed to the revision of the manuscript and approved the final submitted version. 

% \section*{Funding}
% ****** Details of all funding sources should be provided, including grant numbers if applicable. Please ensure to add all necessary funding information, as after publication this is no longer possible.

% \section*{Acknowledgments}
% This is a short text to acknowledge the contributions of specific colleagues, institutions, or agencies that aided the efforts of the authors.

\bibliographystyle{frontiersinSCNS_ENG_HUMS}
\bibliography{mybib}

\end{document}